\documentclass[runningheads]{llncs}

\usepackage{amssymb}
\usepackage{amsmath}
\usepackage{url}
\usepackage{graphicx}
\usepackage{multirow}
\usepackage{wrapfig}
\usepackage{hyperref}
\usepackage{color}
\usepackage{graphicx}
\usepackage[font=small,labelfont=bf]{caption}
\usepackage{enumitem}
\usepackage{multicol}
\usepackage{bbm}
\usepackage{calrsfs}
\usepackage{booktabs}
\usepackage{stmaryrd}
\usepackage{CJK}
\DeclareMathAlphabet{\pazocal}{OMS}{zplm}{m}{n}
\begin{document}

\title{UXNet: Searching Multi-level Feature Aggregation for 3D Medical Image Segmentation}

\author{Yuanfeng Ji\inst{1,2}\and 
	Ruimao Zhang\inst{2}\and
	Zhen Li\inst{3} \and
	Jiamin Ren\inst{2} \and \\
	Shaoting Zhang\inst{2} \and
	Ping Luo\inst{1}\thanks{Ping Luo is the corresponding author of this paper.}}

\authorrunning{F. Author et al.}
\institute{The University of Hong Kong \and
	SenseTime Research \and
	Shenzhen Research Institute of Big Data, The Chinese University of Hong Kong, Shenzhen, Guangdong, China
	 \\}

\titlerunning{UXNet: Searching Multi-level Feature Aggregation for Segmentation}
\maketitle
\begin{abstract}
Aggregating multi-level feature representation plays a critical role in achieving robust volumetric medical image segmentation, which is important for the auxiliary diagnosis and treatment.
Unlike the recent neural architecture search (NAS) methods that typically searched the optimal operators in each network layer,
but missed a good strategy to search for feature aggregations,
this paper proposes a novel NAS method for 3D medical image segmentation, named UXNet, 
which searches both the scale-wise feature aggregation strategies as well as the block-wise operators in the encoder-decoder network. 
UXNet has several appealing benefits.
(1) It significantly improves  flexibility of the classical UNet architecture, which only aggregates feature representations of encoder and decoder in equivalent resolution.
(2) A continuous relaxation of UXNet is carefully designed, enabling its searching scheme performed in an efficient differentiable manner. 
(3) 
Extensive experiments  demonstrate the effectiveness of UXNet compared with recent  NAS methods for medical image segmentation.
The architecture discovered by UXNet outperforms existing state-of-the-art models in terms of Dice on several public 3D medical image segmentation benchmarks, especially for the boundary locations and tiny tissues.
The searching computational complexity of UXNet is cheap, enabling to search a network with best performance less than 1.5 days on two TitanXP GPUs.



	
\end{abstract} 
\section{Introduction}
\label{sec:intro}
\vspace{-0.3cm}

Volumetric medical image segmentation, which provides the detailed pixel-wise categorization of organ regions, is critical to a series of medical analysis, e.g., lung tumour detection~\cite{chlebus2018automatic,li2018h}, gland disease classification~\cite{kirschner2012automatic,yu2017volumetric}. 
Recently, a family of deep models, including 
fully convolutional networks (FCNs)~\cite{long2015fully} and 3D convolutions~\cite{tran2015learning},
has been proposed to improve the segmentation accuracy, 
by extracting the powerful feature representation of organ regions.
%
%
However, a lot of the progress of deep models come from the design of neural network architectures, which heavily relies on expert domain knowledge.

Inspired by the AutoML~\cite{cai2018proxylessnas,liu2018darts}, there has been significant interest in automatically searching the neural network architecture (NAS) through the given searching space.
The goal of NAS is to discover better neural network architectures with the higher performance, the fewer parameters, and even lower computation cost~\cite{cai2018proxylessnas,liu2018darts}.
For medical image segmentation,  Weng et al.\cite{weng2019unet}, Kim et al. \cite{kim2019scalable}, Bae et al. \cite{bae2019resource}, Zhu et al. \cite{zhu2019v} explore to search the building blocks to construct UNet\cite{ronneberger2015u} structure in a gradient-based manner or reinforce learning methods. 
Yu et al. \cite{yu2019c2fnas} further develop a more effective search strategy to alliterative the huge memory-cost problems caused by the 3D task.
In the field of computer vision, Liu et al. \cite{liu2019auto} also search resolution sampling strategy, which is a operator configuration.
Despite the successes of these methods, the searching schemes mainly focus on \emph{searching the effective operators in different layers}.
However, the huge variations in abnormalities’ size, shape, location in 3D medical images usually require information from multi-level feature representations for the robust and dense prediction (\emph{i.e.} multi-level feature aggregations).



\begin{figure}[t!]
\setlength{\abovecaptionskip}{0.cm}
\setlength{\belowcaptionskip}{-5 mm}
	\centering
	\includegraphics[width=0.9\linewidth]{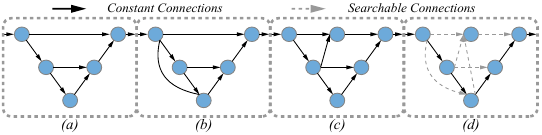}
	\caption{\textbf{Comparisons} of UXNet with previous methods. UXNet's feature aggregation search space in (d) is more general than existing manual designs for medical image segmentation such as (a) UNet (Olaf et al. \cite{ronneberger2015u}), (b) Res/Dense-Unet (Li et al. \cite{li2018h}, Yu et al. \cite{yu2017volumetric}), and (c) Deep Layer Aggregation (Zhou et al. \cite{zhou2018unet++}).}
	\label{fig: diff_arch}
\end{figure}
In literature, many previous studies have demonstrated that aggregation of multi-level features could address the issues of the huge variations in abnormalities for more accurate segmentation~\cite{ibtehaz2020multiresunet,li2018h,lin2018scn,ronneberger2015u,zhou2018unet++}.
Intuitively, merging the high-level and low-level features extracted from different layers helps to enrich semantic representation and capture detailed information.
For example, the latest studies~\cite{ibtehaz2020multiresunet,isensee2018nnu,li2018h,zhang2019progressively} designed to propagate coarse semantic context information back to the shallow layers through top-down and lateral skip-connections, where different layers have various size of 3D receptive field capturing the multi-scale context. 
More recently, Zhou et al.~\cite{zhou2018unet++} exploit more effectively connected architectures via deeper aggregation strategy, which iteratively and hierarchically merges features across adjacent layers, yielding better performance.
However, all of the existing aggregation strategies under the preset are \emph{designed manually}.
Specifically, they perform feature fusion among adjacent or all layers based on a fixed pattern, which may miss useful or involve useless information.

\begin{figure}[t!]
	\centering
	\includegraphics[width=0.9\linewidth]{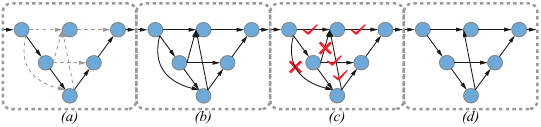}
	\caption{An illustration of our multi-scale search architecture, including (a) all candidate
		searchable connections (grey dashed arrows) are initially regarded as equal, (b) joint optimization of the architecture weights (\emph{i.e.} operators and feature aggregations) and the network weights by solving a bi-level optimization, (c) pruning the learned network according to some preset rules, and (d) the final searching result.}
	\label{fig: diff_arch_multiscale}
	\vspace{-0.5cm}
\end{figure}

To address the above drawbacks, 
we advocate the idea that searching block-wise operators  as well as scale-wise aggregations strategies are equally important for medical segmentation task,
and investigate a novel searching method, UXNet, for more general aggregation search space as shown in Fig.~\ref{fig: diff_arch}.
Concretely, during the searching process, UXNet allows each layer of the network to select optimum operation (\emph{e.g.} traditional convolution or dilated convolution) with a \emph{proper receptive field} to generate better feature representations. 
As shown in Fig.~\ref{fig: diff_arch_multiscale}, based on the extracted multi-level feature representations, a searching strategy for multi-level feature aggregation is further conducted to discover a more efficient fusion method (i.e. which levels of the feature representations are selected to be aggregated in a specific node) for precise segmentation. 
Besides, the block-wise operators and scale-wise aggregations can be jointly searched in a differentiable manner through a continuous relaxation. 
Thus, the overall optimization process of NAS can be automatically driven by the segmentation accuracy,
surpassing the methods of using a pre-fixed set of 3D receptive fields to construct the multi-scale context modelling. 

The \textbf{main contributions} of this paper are three folds.
(1) We present a novel architecture searching setting: searching for the optimal multi-level feature aggregation strategy to fuse the feature maps in UNet-like architecture for 3D medical image analysis.
(2) A novel UXNet searching scheme is proposed by leveraging block-wise operation searching, as well as scale-wise aggregation searching in a uniform framework.
(3) Extensive experiments demonstrate that UXNet outperforms existing state-of-the-art results on most challenging semantic segmentation benchmarks on the 3D Medical Segmentation Decathlon (MSD) challenge \cite{simpson2019large}. UXNet's computational complexity is cheap, such that the best-performing network can be searched less than 1.5 days on two TitanXP GPUs. 
%

\section{Methods}
\begin{figure}[t!]
\textbf{\setlength{\abovecaptionskip}{0.cm}
\setlength{\belowcaptionskip}{-4mm}}
	\centering
	\includegraphics[width=0.95\linewidth]{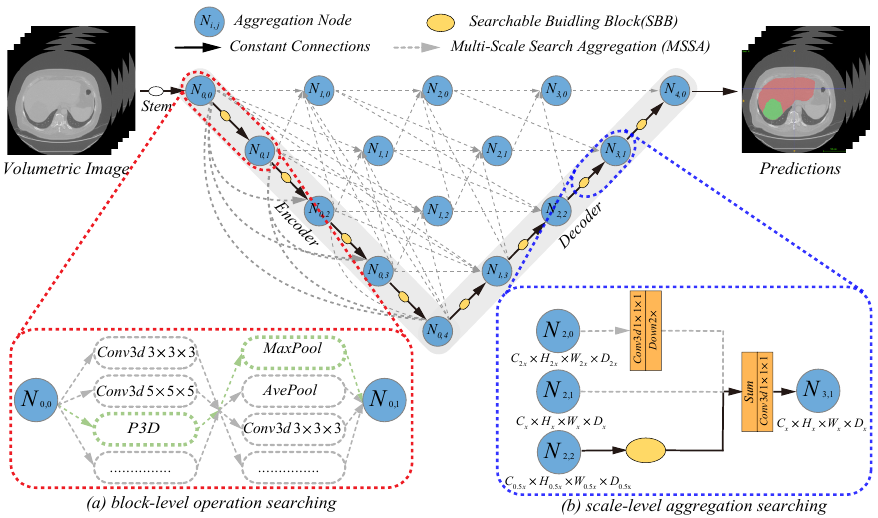}
	\caption{The overall searching space of UXNet, our goal is to implement scale-wise aggregation searching over multi-stage to learn effective feature aggregation for precise segmentation in (b). Besides, the block-wise operation searching is also conducted to search optimum operations for each layer in (a). Blue circle and yellow oval represent the feature aggregation node and the network building block respectively.}
	\label{fig:overview}
	\vspace{-0.20in}
\end{figure}

\vspace{-0.2cm}
We illustrate the proposed UNXet in Fig.~\ref{fig:overview}. The network has encoder and decoder architectures, which is the same as classical UNet, along with the \emph{Searchable Building Block} (SBB) and \emph{Multi-Scale Searchable Aggregation} (MSSA) architecture in-between. 
The former is applied to search the optimum operations in each layer, while the later is used to determine whether or not aggregate the feature maps from various levels in each node.

In practice, we input the volumetric image into the encoder network, producing convolution feature maps (i.e., $N_{0,0}, N_{0,1}, N_{0,2}, N_{0,3}, N_{0,4}$) at different levels. For each levels, input feature maps are fed into a SBB, which enables a flexible combination of various convolution and pooling operators (i.e. the yellow ellipse in Fig.~\ref{fig:overview}), to do the transformation.
%
%
%

MSSA aggregates multi-scale information for assisting the segmentation of organ regions having various sizes. 
As illustrated in Fig.~\ref{fig:overview}, MSSA includes several stages. 
At each stage, the feature maps from all of the levels are firstly regarded as the candidates for aggregation to generate feature maps of the next stage (e.g., at the $0^{th}$ stage, $N_{0,0}, N_{0,1}, N_{0,2}, N_{0,3}, N_{0,4}$ is connected to $N_{1,1}$ at the $1^{st}$ stage). 
It is worthy note that the encoder network also involves candidate dense connections for feature aggregation. 
Compared with the existing approaches that adjust the weight of the connection, MSSA further enables/disables the connection based on its importance to the final recondition task. 
It facilitates a more straightforward way to guide the search of a connection between feature maps at different stages. 
Besides, thanks to SBBs that preserve useful information, MSSA can simplify the search process by eliminating the unnecessary lowest-resolution feature maps at each stage (e.g., $N_{1,4}$ and $N_{2,3}$), while yielding better segmentation result.

\subsection{Searchable Building Block}
\vspace{-0.2cm}
As illustrated in Fig.~\ref{fig:overview}(a),
SBB takes input as a conventional feature map. The block has two layers of convolution/pooling operations. In each layer, we search for an appropriate operation from all options (e.g. traditional convolution or dilated convolution),  using the appropriate receptive field to capture useful image content at different levels.

In each SBB, there are three types of layers: (1) normal layer (i.e., 3$\times$3$\times3$ conv, 3$\times$3$\times$1 conv, 5$\times$5$\times$5 conv, pseudo-3d conv, 2*3$\times$3$\times$3 conv, 3$\times$3$\times$3 conv with rate 2 or 5$\times$5$\times$5 conv with rate 2; (2) reduction layer (i.e., max pooling, average pooling or 3$\times$3$\times$3 conv with stride 2; (3) expansion layer (i.e., transpose conv or trilinear interpolation).
We define the SBB in encoder to be the combination of normal and reduction layers for yielding higher level feature maps. The SBB in the decoder has normal and expansion layers to recover spatial resolution of feature maps.

Specifically, in each layer of SBB, we select from all candidate operations. We formulate this selection as a search process: $y=\sum_{i=1}^{n}\alpha_i O_i(x)$,
where $\alpha_i$ is a learnable weight. Given a layer, we denote $O$ as a set of candidate operations. $x$ is the input feature map to each candidate. Here, the feature map $y$ plays as a weighted summation of outputs of all candidates. During the network training, we optimize learnable weights by using the continuous relaxation~\cite{cai2018proxylessnas}, where the operation having the maximum weight is selected.

\subsection{Multi-Scale Search Aggregation}
\label{sec:MSSA}
\vspace{-0.2cm}

We propose MSSA to learn to connect/disconnect the information propagation pathways. 
As illustrated in Fig.~\ref{fig:overview}, we search  aggregations over multiple  stages. 
At each stage, all levels of feature maps have searchable connections (see the feature aggregation module in Fig.~\ref{fig:overview}(b)) with feature maps at the next stage. 
To compute the feature map $N_{i,j}$ at the $i^{th}$ stage of the $j^{th}$ level, we use searchable connections to combine the multi-scale information as:
\begin{equation}
N_{i,j}=
\begin{cases}
\sum_{k=0}^{j-1} (\sigma(\beta_{i,k\rightarrow i,j }) T(N_{i,k}))  & i=0 \\
\sum_{k=0}^{L-1} (\sigma(\beta_{i-1,k\rightarrow i,j }) T(N_{i-1,k})) & i>0 \\
\end{cases}
\label{eq2}
\end{equation}
where the weight $\beta$ is  learnable weight of each searchable connection, we use $\sigma$ function to map it to $[0,1]$,  which is used to measure the importance of the propagated feature,  and more larger score means the corresponding connection is more important.  
The $i=0$ indicates the aggregations are occurred in the encoder network, and $L$ represents the number of candidate features in last stage. 
The $T$ means a series of transformation to align feature maps with different level, we use a sum operator to aggregate all input features. Fig.~\ref{fig: diff_arch} shows that our multi-scale search space is able to cover existing human-designed networks for medical segmentation.

Furthermore, we minimize the discretization gap with a simple regulation to improve the sparsity, pushing the connection weight to the extreme (i.e., 0 or 1). 
That is, we aim to either enable or disable a specific pathway. The regulation is formulated as:

\begin{equation}
\pazocal{J} = -(\sigma(\beta)-0.5)^2
\label{eq3}
\end{equation}



After searching the connections for aggregating multi-scale information, we prune the searchable pathway to satisfy the rules as follows (as illustrated in Figure~\ref{fig: diff_arch}). 
(a)  A threshold $\tau$ is pre-defined, when $\sigma(\beta)\geq\tau$, we keep the corresponding connections. %
(b) Once the aggregation node has no connection with features in the next stage, we delete the relevant connections.


\begin{figure}[t!]
	\centering \includegraphics[width=0.8\linewidth]{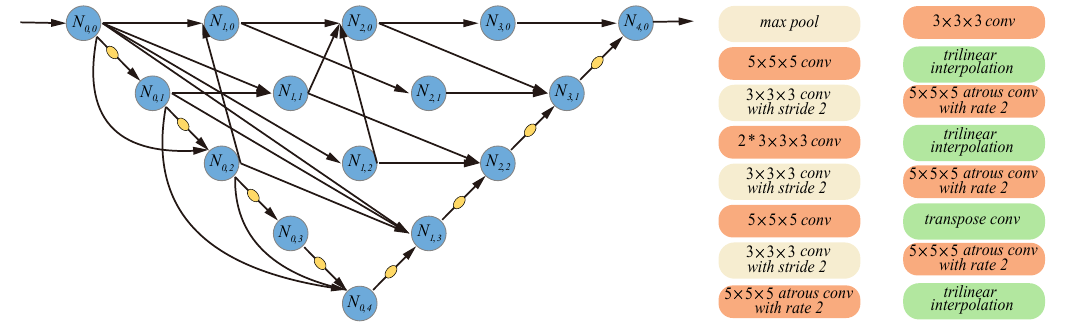}
	\caption{The network discovered by our UXNet on the brain dataset of MSD challenge.}
	\label{fig:search_result}
	\vspace{-0.5cm}
\end{figure}

\subsection{Network Training}
\label{sec:network_training}
\vspace{-0.2cm}
All of the learnable weights (i.e., $\alpha$ in SBB and $\beta$ in MSSA)  can be optimized efficiently using gradient descent. 
To further avoid overfitting the training data, we follow the bi-level optimization policy in \cite{cai2018proxylessnas,liu2018darts}. 
That is, we divide the training data into two sets $trainA$ and $trainB$, which are used to train the set of operation weights $w$ (e.g., convolutional layers) and the set of connection weights $\alpha,\beta$, respectively.
We solve the objective $\pazocal{L}_{trainA}(w, \alpha, \beta)$ for $w$, by fixing the connection weights. Then we fix the operation weights and solve the objective $\pazocal{L}_{trainB}(w, \alpha, \beta)$ and $\pazocal{J}_{trainB}(\beta)$ for the connection weights. 
Here, $\pazocal{L}_{trainA}(w, \alpha, \beta)$ represents the cross-entropy loss. $\pazocal{L}_{trainB}(w, \alpha, \beta)$ and $\pazocal{J}_{trainB}(\beta)$ denotes the L2-regulation (see Eqn.\ref{eq3}).

\section{Experiments}
\vspace{-0.2cm}
\subsection{Datasets and Settings}
\vspace{-0.2cm}

Following previous NAS-based methods, we evaluate  UXNet on three subset of 3D Medical Segmentation Decathlon (MSD) challeng\cite{simpson2019large} (i.e. the brain, heart, and prostate), which contains 484, 20, 32 cases for  training respectively. 
We adopt the same image pre-processing strategy in \cite{isensee2018nnu}.
%
%
Since the annotation of test datasets are not publicly available, we report the 5-fold cross-validation results as in \cite{bae2019resource,isensee2018nnu,kim2019scalable}. 
%
%
We also report the validation results on the 2D lesion segmentation dataset released by the Skin Lesion Segmentation and Classification 2018 challenge\cite{hardie2018skin}, which provides 2594 training images. 
We publish all results in terms of the dice coefficient and the higher score indicates the better result.

%
%
%
%
%

%
When learning network weights $w$ in Sec.~\ref{sec:network_training}, we use Adam optimizer with an initial learning rate of 0.0003, and the betas range from 0.9 to 0.99.
The initial values of $\alpha$ and $\beta$ are set as 1 and 0.  
They are optimized using Adam optimizer with a learning rate of 0.003 and weight. When we finish searching and pruning the network, we retrain the derived network from scratch. The computation of UXNet is cheap by training 1.5 days on two TitanXP GPUs for brain task, which is cheaper than RONASMIS \cite{bae2019resource} that trained 3.1 days on one RTX 2080Ti GPU and SCNAS \cite{kim2019scalable} that trained one day on four V100 GPUs.
Please refer to appendix for more training details of each dataset.

\subsection{Ablation Studies}
\vspace{-0.2cm}
We first conduct the ablation studies on the brain dataset of MSD challenge by removing the critical component of UXNet, i,e., SBB and MSSA. 
In such case, the model degrades to the original UNnet and achieves the score of $72.5$\%. It lags far behind our full model in terms of segmentation performance. 
As shown in Table~\ref{tab:ablation_study}, by adding the SBBs, the performance is promoted to $73.43$\%.
It demonstrates that the SBB enables each layer of the network to produce rich presentation via optimum operation.
Note that we adopt the UNet-Style connections to fuse the features in encoder and decoder network in this case.
When the MSSA is enabled, the discovered optimal feature aggregation strategy could further improve the score  to $74.57$\%.





We further evaluate the influence of hyper-parameter $\tau$ in Sec.~\ref{sec:MSSA}. The higher value indicates the more sparse aggregations.
Table.\ref{tab:ablation_study} shows the results of using different $\tau$ to prune the over-connections. 
we find that using a too low or high threshold to clip the connections will decrease the model's performance.
It indirectly illustrates that involving useless or missing useful information will lead to the poor performance.

\begin{figure}[t!]
	\centering
	\includegraphics[width=0.8\linewidth]{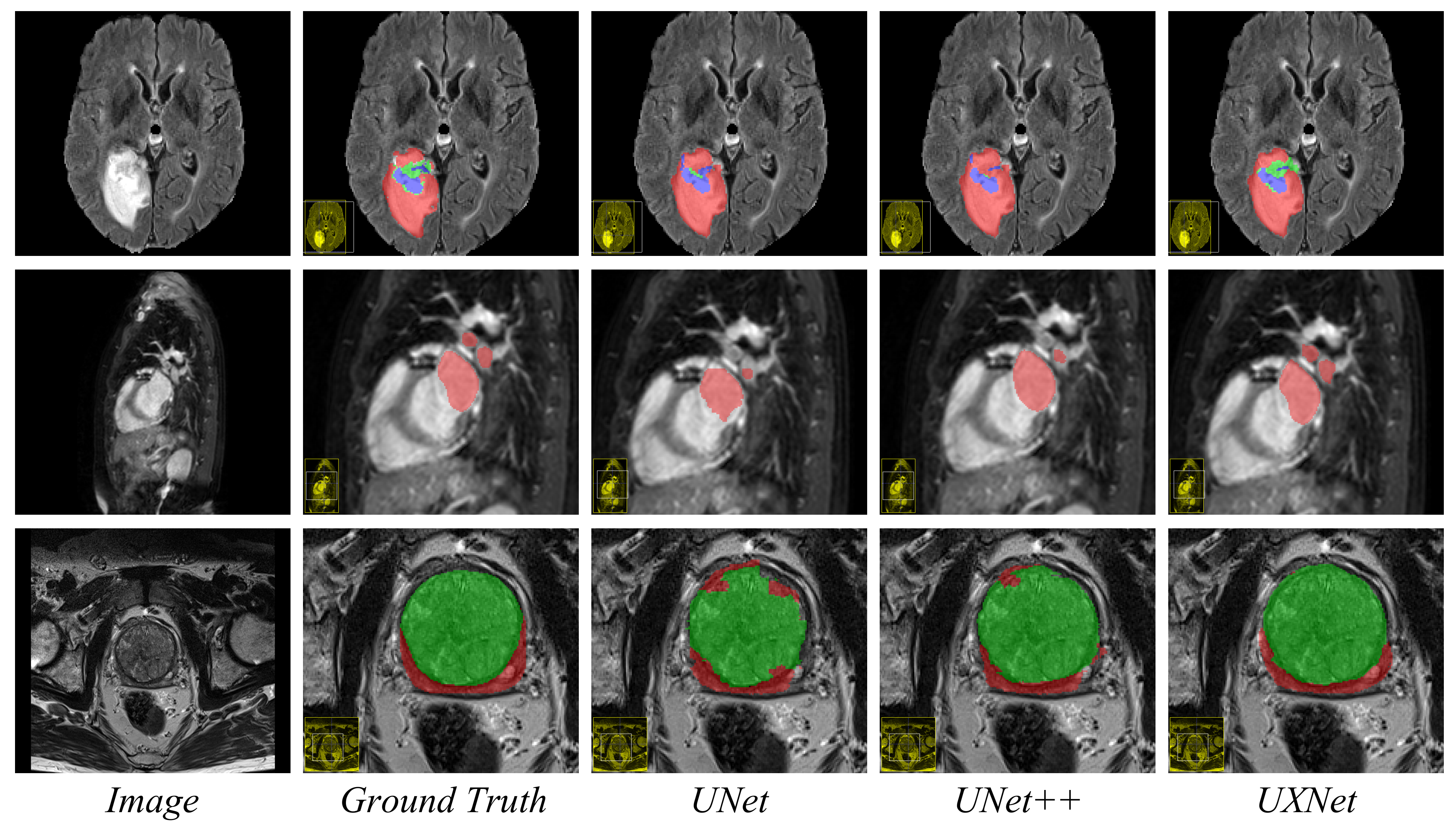}
    \vspace{-0.3cm}
	\caption{The segmentation results of the UNet \cite{ronneberger2015u}, UNet++ \cite{zhou2018unet++} and our UXNet on some challenging cases. Our method achieves more accurate segmentation results.}
	\label{fig:compared_with_sota}
\end{figure}


\begin{table}[t]
    \centering
	{\footnotesize
	{\tabcolsep12pt\def\arraystretch{0.8}
        \vspace{-0.3cm}
		\begin{tabular}{cccc}
			\toprule[1.5pt]
			SBB & MSSA & $\scalebox{1.44}{$\tau$}$ & Dice(\%) \\ 
			\midrule[1.5pt]
			$\texttimes$ & $\texttimes$ & $\textunderscore$ & 72.50  \\
			$\checkmark$ & $\texttimes$ & $\textunderscore$ & 73.43  \\
			$\texttimes$ & $\checkmark$ & 0.75 & 73.65  \\
			$\checkmark$ & $\checkmark$ & 0.60 & 73.90  \\
			$\checkmark$ & $\checkmark$ & 0.75 & \textbf{74.57}  \\
			$\checkmark$ & $\checkmark$ & 0.90 & 74.35  \\
			\bottomrule[1.5pt]
        \end{tabular}}}
        \caption{Ablations studies of different setting on the brain dataset of MSD.}
    \label{tab:ablation_study}
     \vspace{-0.95cm}
\end{table}

\subsection{Comparison with State-of-the-Arts}
\vspace{-0.2cm}

\textbf{Medical Segmentation Decathlon.} The MSD challenge contains ten tasks totally, which provide the multimodal images with a vast variant of shapes and locations. 
In Table.~\ref{tab:compare_with_sota}, we compare our UXNet with state-of-the-art methods in terms of cross-validation results on the brain, heart prostate dataset as in \cite{bae2019resource,isensee2018nnu,kim2019scalable}.
We divide the compared methods into two groups. In this first group, the methods adopt human-design architecture for segmentation. Compared to this kind of methods, our UXNet achieves significant improvement by searching a optimal architecture automatically.
%
%
Some visualization comparisons are also available in Fig.~\ref{fig:compared_with_sota}. 
In this second group, the methods also apply the NAS technique to search architecture.
They mainly focus on designing various searching strategies by using reinforcement learning or differentiable manner but not the task-specific searching space design.
Thus, these kinds of methods did not consider the feature aggregation strategies into search space, limited the flexibility of the models to recognize the organs with narrow areas compared with proposed UXNet.
For a fair comparison, we report the result of UXNet by using a single model with basic data augmentation(i.e., sliding window, flipping). 
The searched architecture still outperforms these state-of-the-art NAS-based methods.
The Fig. \ref{fig:search_result} visualizes the best architecture found on the brain dataset. 
We observed that the atrous convolution with the large kennel is heavily used, suggesting the importance of learning with the large receptive field. 
It can also be seen that the low-level feature is more used compared to other methods, which indicate the importance of spatial details for medical segmentation.

\begin{table}[!btp]
\setlength{\abovecaptionskip}{0.cm}
\setlength{\belowcaptionskip}{-6 mm}
	\centering
	{\scriptsize
		{\tabcolsep=1.5pt\def\arraystretch{1.35}
			\begin{tabular}[width=\linewidth]{c|cc|cccc|c}
				\toprule[1.5pt]
				\multirow{2}{*}{Model}&
				\multicolumn{2}{c|}{Auto Search}&
				\multicolumn{4}{c|}{MSD\cite{simpson2019large}} & \multicolumn{1}{c}{ISIC\cite{simpson2019large}} \\
				& Block & Aggregation & Brain Tumor & Heart & Prostate & Average & Lesion   \\
				\midrule[1.5pt]
				Unet\cite{cciccek20163d} & $\texttimes$& $\texttimes$  &72.5 & 90.70 & 73.13 &78.77 &86.2     \\
				NNUnet\cite{isensee2018nnu} & $\texttimes$& $\texttimes$ &74.00  &92.70 &74.54 &80.41 &\textunderscore  \\ 
				Unet++\cite{zhou2018unet++} & $\texttimes$& $\texttimes$ &72.66 &91.56 &72.95 &79.05 &86.7  \\
				U-ResNet\cite{kim2019scalable} & $\texttimes$& $\texttimes$&71.61 &89.60 &63.77 &75.00 & 88.0     \\ \hline
				SCNAS \cite{kim2019scalable} & $\checkmark$& $\texttimes$ &72.04  &90.47 &67.92 &76.81 &\textunderscore    \\
				RONASMIS \cite{bae2019resource} & $\checkmark$& $\texttimes$ &74.14  &92.72  &75.71 & 80.85 &\textunderscore    \\ \hline\hline
				UXNet($\tau=0.75$) & $\checkmark$& $\checkmark$ &\textbf{74.57}  &\textbf{93.50} & \textbf{76.36} & \textbf{81.48} & \textbf{89.6} \\ 
				\bottomrule[1.5pt]
	\end{tabular}}}
	\caption{Comparison with different approaches on the MSD and ISIC dataset}
	\label{tab:compare_with_sota}
\end{table}

\textbf{Skin Lesion Segmentation and Classification.} We also evaluate our UXNet on the 2D medical image segmentation. 
In Table.\ref{tab:compare_with_sota}, the results of our method still outperforms other conventional baselines by a significant margin.
Such results demonstrate the versatility of proposed UXNet on various segmentation tasks.


\vspace{-0.2cm}
\section{Conclusion}
\vspace{-0.2cm}
In this paper, we propose a general framework of neural architecture search for 3D medical image segmentation, termed UXNet, which searches  scale-wise  feature  aggregation  strategies  as well as  block-wise  operators  in  the  encoder-decoder path.
The careful designed searching space achieves the robust segmentation results.
In addition, the discovered  segmentation architecture reveals the properties of information propagation for the specific dataset.
The further will explore UXNet under the real-word hardware constraints, such as memory,  speed, and power consumption. 
In addition, the task-oriented searching space design will also be a potential research direction.

\section*{Acknowledgments}
This work was supported in part by the Key Area R\&D Program of Guangdong Province with grant No. 2018B030338001, by the National Key R\&D Program of China with grant No. 2018YFB1800800, by Natural Science Foundation of China with grant NSFC-61629101, by Guangdong Zhujiang Project No. 2017ZT07X152,  by Shenzhen Key Lab Fund No. ZDSYS201707251409055, by Open Research Fund
from Shenzhen Research Institute of Big Data No. 2019ORF01005, by NSFC-Youth 61902335, and by the General Research Fund No.27208720.

\bibliographystyle{splncs04}
\bibliography{library}

\end{document}